# Categories of Emotion names in Web retrieved texts

S. Petrov, J.F. Fontanari, and L.I. Perlovsky

*Abstract*— The categorization of emotion names, i.e., the grouping of emotion words that have similar emotional connotations together, is a key tool of Social Psychology used to explore people's knowledge about emotions. Without exception, the studies following that research line were based on the gauging of the perceived similarity between emotion names by the participants of the experiments. Here we propose and examine a new approach to study the categories of emotion names - the similarities between target emotion names are obtained by comparing the contexts in which they appear in texts retrieved from the World Wide Web. This comparison does not account for any explicit semantic information; it simply counts the number of common words or lexical items used in the contexts. This procedure allows us to write the entries of the similarity matrix as dot products in a linear vector space of contexts. The properties of this matrix were then explored using Multidimensional Scaling Analysis and Hierarchical Clustering. Our main findings, namely, the underlying dimension of the emotion space and the categories of emotion names, were consistent with those based on people's judgments of emotion names similarities.

## I. INTRODUCTION

The concept of emotion has been used in various ways by different authors, sometimes denoting different physiological or mental mechanisms, and often without any attempt to clarify its intended meaning [1]. There were a variety of attempts to define emotion in the many disciplines where the concept is relevant. For instance, according to Grossberg and Levine, emotions are neural signals indicating satisfaction or dissatisfaction of instinctual needs [2] – not unlike Simon's view that emotions serve to interrupt normal cognition when unattended goals require servicing [3]. Emotions have been related to survival [4] and to facial expressions [5]. Emotions have also been tied to social interactions, arising from the dissonance people feel between competing goals and conflicting interpretations of the world, a view which has been subsumed in the belief–desire theory of emotion [6]. Cabanac defined emotions as any mental experience with high intensity and high hedonic content [7]. Most authors seem to agree on emotions performing appraisals of situations or events, though this point has been proved only in the beginning of the 1990s through the study of damaged prefrontal cortex patients [8]. In fact, appraisal theory, whose tenet is that emotions depend on perceiving the relational meaning of encounters with the environment, is probably the most popular psychological account of emotion nowadays [9].

Another controversial issue related to the concept of emotion is the attempt by some authors to define the so-called basic emotions. In fact, although it is in general agreed that basic emotions evolved with fundamental life tasks [10], [11], [12], [13], [14], there are about 14 different proposals of emotion candidates for this primary category, whose size can vary from 2 to 11 members [15]. Similarly to the role of primary colors in vision, all other (non-basic) emotions could be thought of as a composition of a few basic emotions [4], [14]. The idea of basic emotions having specialized neurophysiological and anatomical substrates or being the primitive building blocks of other, non-basic, emotions has been criticized [15], but there are arguments in support of this idea based on the interactions between emotions and cognition [16]. Some cognitive scientists use the name 'discrete' instead of basic [1], [17]. Grossberg and Levine theory relating emotions to instincts [2] could have been used for relating basic emotions to basic (or bodily) instincts; this direction of research was not pursued to our knowledge. Perlovsky argued that human 'higher' emotions, such as aesthetic, musical, and cognitive dissonance emotions are related to the instinct for knowledge, a 'higher' instinct; that these emotions are principally different from basic emotions, and their number is much larger being better described by a continuum of emotions rather than by discrete labels [18], [19], [20]. Steps toward experimental test of this hypothesis were made in [21] and [22].

Regardless of the experts' theories and disputes on emotions, people have an informal and implicit naïve theory of emotion which they use in their daily routines to interact and influence other individuals. Emotion words are labels for the categories of the folk taxonomy of emotional states, and have an immense importance in clinical and personality psychological evaluations which use mood adjective (emotion name) checklists to assess the patients' emotional states [23]. Hence, the understanding of folk emotion categories has an interest and importance on its own. In addition, in an intuitive leap one may associate properties of emotion word categories with those of 'true' psychological emotions. Such a relationship is a hypothesis, which may or may not be correct for various purposes and establishing limits of validity of this relationship is a separate scientific problem, which we will not address in this contribution.

The research at São Carlos was supported by The Southern Office of Aerospace Research and Development (SOARD), grant FA9550-10-1-0006, and Conselho Nacional de Desenvolvimento Científico e Tecnológico (CNPq).

S. Petrov is an independent consultant at 10 Dana Street # 508 Cambridge, MA 02138 (e-mail: yukpun60@gmail.com).
J. F. Fontanari is with Instituto de Física de São Carlos, Universidade de São Paulo, São Carlos, SP, Brazil (phone: +55-16-33739849; fax: +55-16-33739877; e-mail: fontanari@ifsc.usp.br).
L. I. Perlovsky is with Harvard University, Cambridge MA and The Air Force Research Laboratory, Wright-Patterson Air Force Base, OH (e-mail: Leonid.Perlovsky@wpafb.af.mil).

Nevertheless, as the manner humans perceive color similarities can tell much about the physical distance (in terms of the wavelengths) between the colors [24], [25] it is not so far-fetched to expect that the way people think and talk about emotions may bear some relationship to psychological emotional states.

Most quantitative approaches to understanding the underlying organization of the emotion categories have focused on perceived similarities among (English) emotion names. A remarkable outcome of this research avenue was the finding that emotion labels are not independent of each other [23]. Attempts to produce a structural model to capture the relationships among the emotion word categories led to the proposal of the Circumplex model in which emotion names are arranged in a circular form with two bipolar dimensions interpreted as the degree of pleasure and the degree of arousal [26], [27]. In that sense, emotion names mix together in a continuous manner like hues around the color circle [23]. A reorientation and, consequently, reinterpretation of the axes of the Circumplex model as positive affect and negative affect has been suggested to correct for the fact that there were few emotions in the neutral middle region of the pleasantness-unpleasantness axis [28].

A complementary approach to the structural models of emotion names categories, such as the Circumplex model, is the exploration of the hierarchical structure of those categories [29]. This more intuitive approach allows the immediate identification of the basic emotions categories as those that are closer to the root of the hierarchical tree. Both approaches are characterized by the use of non-elementary mathematical techniques (e.g., factor analysis, multidimensional scaling and hierarchical clustering) to extract relevant information from a similarity matrix produced by asking individuals to rate the similarity between a given set of distinct emotion words.

The procedure to obtain the similarities among the emotion words is the main feature that distinguishes our contribution from the landmark papers mentioned in the previous paragraphs. Rather than asking individuals to rate the similarities using a fixed discrete scale, we search for texts in the Web that contain emotion names and define the similarity among a specified set of target emotion names - essentially the same set used in the study of [29] – as the number of common words in the close neighborhoods (contexts) of the target emotion names. This definition allows us to express the similarities as dot-products between vectors in the space of contexts and then use the full power of linear algebra for its analysis. In particular, we follow the original multidimensional scaling framework [30] and re-express these similarities as dot-products of orthogonal vectors in the space spanned by the target emotion names. These vectors, known as principal coordinates, are nothing but the rescaled eigenvectors of the similarity matrix, whose entries give the numerical values of the similarity between all pairs of target emotion names. The dimension $m^*$ of the emotion space estimated by the multidimensional scaling analysis is not inconsistent with the estimates based on the individuals' judgment of the similarities between emotion names. Given the flexibility of the interpretation of the elbow test, we find that both $m^* \approx 5$ and $m^* \approx 30$ are permissible estimates. Regarding the hierarchical clustering analysis, our clusters exhibit a good correlation to those produced by [29], but also show significant differences as the grouping of antagonistic emotion names together such as *pleasure* and *displeasure*, for example.

The rest of this paper is organized as follows. In Section II we describe the procedure we used to extract and then to clean up texts from the Web. That section also contains our definition of the similarity between pairs of emotion names and its mathematical interpretation as a dot product in the linear vector space of contexts. The resulting similarity matrix $S$ is then analyzed in Section III. We begin with some elementary statistical measures and then proceed to the Multidimensional Scaling Analysis, which involves the spectral decomposition of $S$ [30]. The section ends with the presentation of the categories into which the emotion names are grouped according to Ward's minimum variance hierarchical clustering algorithm [31]. Finally, in Section IV we present our concluding remarks and outline the future research directions.

## II. EXPERIMENTAL METHOD

Practically all methods employed in the literature to investigate the closeness of common emotion names were based on querying participants about the similarity and differences between a given set of emotion names [23], [29] [32], [33]. Our approach departs from the traditional psychology methods in that we gauge the similarity between two emotion names by comparing the contexts in which they are used in documents extracted from the Web. At the present stage, we do not explore the semantic information contained in those texts; rather our comparison is based solely on the shared vocabulary between documents.

### A. Target emotion names

Although contemporary English contains hundreds of terms with emotional connotations [34], apparently there is no consensus on which of these terms can be considered emotion names or emotion prototypes. An ingenious approach to this issue was offered by [29] who presented a list of 213 candidate emotion names to 112 students and asked them to rate those terms on a 4-point scale ranging from 'I definitely would not call this an emotion' to 'I definitely would call this an emotion'. This procedure resulted in a much shorter list containing 135 emotion names that the participants rated highest on the 4-point 'emotionness' scale. In addition to these 135 emotion names we have included 7 more names, namely, *anticipation*, *acceptance*, *wonder*, *interest*, *aversion*, *pain*, and *courage* in order to take into account a few widely recognized 'basic' emotions [15] which were not in the original list of that study. Table I shows the 135 emotion names from the list of [29] together with the 7 names mentioned above, totaling 142 emotion names which we use as target words in our Web queries, as described next.

TABLE I

| | | |
|---|---|---|
| 1 : acceptance | 49 : fondness | 97 : pity |
| 2 : adoration | 50 : fright | 98 : pleasure |
| 3 : affection | 51 : frustration | 99 : pride |
| 4 : aggravation | 52 : fury | 100 : rage |
| 5 : agitation | 53 : gaiety | 101 : rapture |
| 6 : agony | 54 : gladness | 102 : regret |
| 7 : alarm | 55 : glee | 103 : rejection |
| 8 : alienation | 56 : gloom | 104 : relief |
| 9 : amusement | 57 : glumness | 105 : remorse |
| 10 : anger | 58 : grief | 106 : resentment |
| 11 : anguish | 59: grouchiness | 107 : revulsion |
| 12 : annoyance | 60 : grumpiness | 108 : sadness |
| 13 : anticipation | 61 : guilt | 109 : satisfaction |
| 14 : anxiety | 62 : happiness | 110 : scorn |
| 15 : apprehension | 63 : hate | 111 : shame |
| 16 : arousal | 64: homesickness | 112 : shock |
| 17 : attraction | 65 : hope | 113 : sorrow |
| 18 : aversion | 66 : hopelessness | 114 : spite |
| 19 : bitterness | 67 : horror | 115 : suffering |
| 20 : bliss | 68 : hostility | 116 : surprise |
| 21 : compassion | 69 : humiliation | 117 : sympathy |
| 22 : contempt | 70 : hurt | 118 : tenseness |
| 23 : courage | 71 : hysteria | 119 : terror |
| 24 : defeat | 72 : infatuation | 120 : torment |
| 25 : dejection | 73 : insecurity | 121 : triumph |
| 26 : delight | 74 : insult | 122 : uneasiness |
| 27 : depression | 75 : interest | 123 : unhappiness |
| 28 : desire | 76 : irritation | 124 : vengefulness |
| 29 : despair | 77 : isolation | 125 : woe |
| 30 : disappointment | 78 : jealousy | 126 : wonder |
| 31 : disgust | 79 : jolliness | 127 : worry |
| 32 : dislike | 80 : joviality | 128 : wrath |
| 33 : dismay | 81 : joy | 129 : zeal |
| 34 : displeasure | 82 : jubilation | 130 : zest |
| 35 : distress | 83 : liking | 131 : tenderness |
| 36 : dread | 84 : loathing | 132 : thrill |
| 37 : eagerness | 85 : loneliness | 133 : caring |
| 38 : ecstasy | 86 : love | 134:sentimentality |
| 39 : elation | 87 : lust | 135: longing |
| 40 : embarrassment | 88 : melancholy | 136: cheerfulness |
| 41 : enthusiasm | 89 : misery | 137: enjoyment |
| 42 : envy | 90 : mortification | 138: contentment |
| 43 : euphoria | 91 : neglect | 139: enthrallment |
| 44 : exasperation | 92 : optimism | 140: amazement |
| 45 : excitement | 93 : outrage | 141: astonishment |
| 46 : exhilaration | 94 : pain | 142: nervousness |
| 47 : fear | 95 : panic | |
| 48 : ferocity | 96 : passion | |

*B. Context retrieval*

For every target emotion name listed in Table 1, we retrieve 99 documents containing the target word from the Web using the Yahoo! search engine. Thus, the documents were ordered by Yahoo! relevance criteria. Since, as expected, almost every target emotion word is used in a variety of semantic contexts which are unrelated to emotions and many of them appear in advertising, our search focused on documents in which the target emotion word is combined with the word 'emotion'. This combination more or less restricted the retrieved documents to ones where a particular emotion – or at least an emotion word – was the subject of the text. This combination – target emotion name plus the word 'emotion' - increased considerably the average length of the retrieved documents.

These retrieved texts were then cleaned up for the purpose of forming the so called bags of words. A bag of words is a list of words in which the grammatical rules are ignored. During the cleanup, all words of length 2 or shorter were eliminated. In addition, we have also filtered out conjunctions, prepositions, pronouns, numbers, punctuations marks and all formatting signs. In what remained of each document, we then selected a sequence of 41 consecutive words with the target emotion name in the middle, i.e., 20 words before and 20 words after the target word. Only the 50 more relevant (according to Yahoo!) contexts were retained for every emotion name. For some of the emotion names used by [29], namely, *tenderness*, *thrill*, *caring*, *sentimentality*, *longing*, *cheerfulness*, *enjoyment*, *contentment*, *enthrallment*, *amazement*, *astonishment* and *nervousness*, we were unable to retrieve 50 contexts of the prescribed length out of the 99 retrieved ones, and so we excluded those words (numbered 131 to 142 in Table I) from our list of target emotion names. We indexed these 130 emotion words by $i = 1, \cdots, 130$.

In summary, for each of the first 130 target emotion names exhibited in Table I, we produced 50 distinct sequences of words, each containing 20 valid words before and 20 valid words after the target word in question. A valid word is a word that escaped the cleanup procedure applied to the Web documents retrieved by the Yahoo! search engine. The final step is to lump all the 50 sequences corresponding to a given target emotion name, say word *i*, into a single bag of words which we denote by $\Omega_i$. (Note that $\Omega_i$ is not a set since an element can be present there more than once.) Hence, the number of elements in a bag of words is $50 \times (20 + 1 + 20) = 2050$, regardless of the emotion index-name $i = 1, \cdots, 130$. We found only $K = 34244$ distinct words among the 266500 words that make up the 130 word bags.

*C. Similarity measure*

The basic similarity $\hat{S}_{ij}$ between the two target emotion words *i* and *j* is calculated using their corresponding bags of words, $\Omega_i$ and $\Omega_j$, as follows. Let us denote by $\omega_{ij}(k)$ the number of times word $k$ from $\Omega_i$ appears in the bag $\Omega_j$. Note that $\omega_{ij}(k)$ and $\omega_{ji}(k)$ have different domains since there might be words that belong to $\Omega_i$ but not to $\Omega_j$, and vice-versa. The unprocessed similarity $\hat{S}_{ij}$ is defined as

$$\hat{S}_{ij} = \sum_k \omega_{ij}(k) \qquad (1)$$

where $k$ runs over all words (repetitions included) in $\Omega_i$. This procedure takes into account multiple appearances of words in bags $\Omega_i$ and $\Omega_j$. For example, if word $k$ from bag $\Omega_i$ appears $m$ times in bag $\Omega_j$ and $n$ times in bag $\Omega_i$ then it contributes with the factor $nm$ to the unprocessed similarity $\hat{S}_{ij}$. From this example we can easily realize that the similarity measure defined by eq. (1) is symmetric, i.e., $\hat{S}_{ij} = \hat{S}_{ji}$ for all pairs of target emotion words $i$ and $j$. In the case the bags $\Omega_i$ and $\Omega_j$ consist of the same word repeated $n$ times we have $\hat{S}_{ij} = n^2$, whereas if $\Omega_i$ and $\Omega_j$ do not have any element in common we have $\hat{S}_{ij} = 0$.

We note that our definition of the unprocessed similarity, eq. (1), is equivalent to a dot-product of two vectors. In fact, let us order all the $K = 34244$ distinct words alphabetically (the specific order is not essential for the argument). Then we can represent each bag $\Omega_j$ uniquely by a $K$-dimensional vector $Y_j = (Y_j^1, \cdots, Y_j^K)$ in which the component $Y_j^l$ ($l = 1, \cdots, K$) is the number of times that word $l$ appears in $\Omega_j$. Hence our unprocessed similarity measure can be written as the dot-product

$$\hat{S}_{ij} = \sum_{l=1}^{K} Y_i^l Y_j^l. \qquad (2)$$

Of course, the vectors $Y_j$ are very sparse, i.e., most of their components are zeros. We will offer a more economic representation of the similarity entries in terms of a much lower dimension vector space in Section III. Equation (2) is important because it shows that $\hat{S}_{ij}$ is a dot-product which allows us then to use the full power of linear algebra for its analysis. Even more important, however, is the observation that $\hat{S}_{ij}$ provides little information about the proximity or closeness of the vectors $Y_i$ and $Y_j$, unless these vectors are normalized. In fact, one the one hand almost orthogonal vectors can have a very high unprocessed similarity value if their norms are large and, on the other hand, two parallel vectors may have a low similarity if their norms are small. This problem can be easily corrected by defining the normalized similarity as

$$S_{ij} = \frac{\hat{S}_{ij}}{\sqrt{\hat{S}_{ii}\hat{S}_{jj}}} \qquad (3)$$

where $\hat{S}_{ii} = \sum_{l=1}^{K}(Y_i^l)^2$ is the squared norm of the vector $Y_i$. Note that $S_{ij} \in [0,1]$ and $S_{ii} = 1$. Henceforth we will refer to the normalized similarity, eq. (3), simply as the similarity between emotion index-names $i$ and $j$.

*D. Null random model*

Since our approach is based on the statistics of word contexts, we should also define a 'null' model using random contexts, so that we could identify which results depend on contexts specific to emotion words, and which ones characterize random contexts. A null model to compare our results can be obtained as follows. First, we lump together the 130 word bags into a single meta-bag comprising 266500 elements. Next we pick 2050 ($41 \times 50$) words at random and without replacement to form the random bag $\Phi_1$. This drawing step is repeated to form the remaining word bags $\Phi_2 \cdots, \Phi_{130}$, which ends when the meta-bag is emptied. Given these randomly assembled word bags, we then follow the procedure described before to calculate the normalized entries $R_{ij}$ of the random similarity matrix $\mathbf{R}$ between emotion names $i$ and $j$.

III. RESULTS

In this section we use two techniques often employed in the investigation of the underlying psychological structure of the use of emotion words by English speakers, namely, multidimensional scaling and hierarchical clustering analysis [29], [35]. However, before we introduce these more involved exploratory tools, we present a simple statistical description of the $130 \times 130$ similarity matrix $\mathbf{S}$, as well as of its random counterpart $\mathbf{R}$, generated according to the procedure described in the previous section.

*A. Simple Statistical Measures*

We begin with the most elementary measures, namely, the mean $\bar{S}$ and the standard deviation $\sigma_S$ of the entries of the similarity matrix $\mathbf{S}$. We find $\bar{S} = 0.376$ and $\sigma_S = 0.069$. For the random null model, these two quantities are $\bar{R} = 0.686$ and $\sigma_R = 0.020$. The considerable differences between even these simple measures indicate a rich underlying structure of the matrix $\mathbf{S}$. A better visualization of the entries of these matrices is obtained by ordering the entries according to their rank, from the largest to the smallest. Disregarding the diagonal elements, there are $130 \times 129/2 = 8385$ distinct entries in each of these matrices and in Fig. 1 we present their values as function of their ranks $r = 1,\cdots,8385$. These distributions are remarkable symmetric around their mean values, shown by the horizontal lines in the figure. For the most part of the rank order range, say $2000 < r < 7000$, the similarity values decrease linearly with the rank $r$. In particular, in this range we find the following fittings $S_r \approx 0.463 - 2.1 \times 10^{-5} r$ and $R_r \approx 0.714 - 6.4 \times 10^{-6} r$. These results indicate that the random similarity matrix $\mathbf{R}$ is much more homogenous than $\mathbf{S}$, which is expected since in the random model the contexts do not provide information to distinguish between the target words.

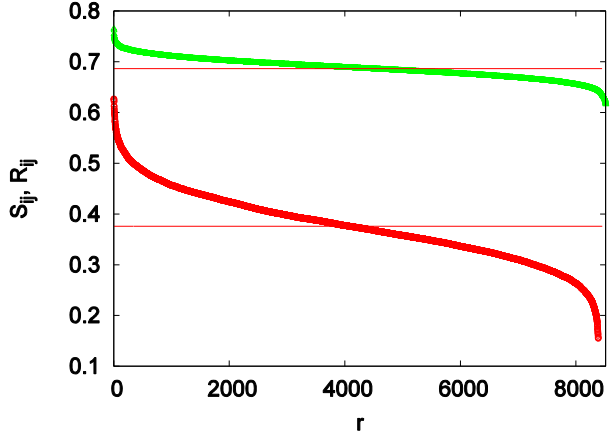

Fig. 1. Plot of the off-diagonal entries of the similarity matrix $S$ (circles, lower curve) and its random version $R$ (triangles, upper curve) as function of their ranks $r = 1, \cdots, 8385$. The horizontal solid lines are the mean values $\bar{S} = 0.376$ and $\bar{R} = 0.686$. These results corroborate the expectation that $R$ is more homogeneous than $S$.

To conclude this section it is instructive to write out the pair of words corresponding to the extremes of the rank order distribution. For instance, the pair of distinct words with the largest similarity, $S_1 = 0.627$, is *aggravation* and *irritation*; the pair with the second largest, $S_2 = 0.626$, is *anguish* and *gloom*; the pair with the third largest, $S_3 = 0.623$, is *mortification* and *shock*; and the pair with the smallest similarity, $S_{8385} = 0.155$, is *hostility* and *glee*. The high similarity of emotion words of close emotional content lends credibility to our experimental procedure.

### B. Multidimensional Scaling

Similarly to most psychological experimental settings aiming at exploring the relationships between $n$ emotion words [23], [29], the end product of our data-mining methodology is a $n \times n$ similarity matrix $S$. It is thus tempting to assume the existence of a subjacent 'emotion' vector space of dimension $m$ that contains vectors whose Euclidian scalar product generates $S$. The mathematical procedure to derive a base of this vector space is known as Multidimensional Scaling Analysis [30], [35], [36]. More explicitly, we want to find the set of $m$ orthogonal vectors of length $n$, $(x_1^a, x_2^a, \cdots, x_n^a)$ with $a = 1, \cdots, m$, such that

$$S_{ij} = \sum_{a=1}^{m} x_i^a x_j^a \quad (4)$$

for all pairs $i, j = 1, \cdots, n$. This problem has a simple and neat solution in the case $m = n$. In fact, denoting the eigenvectors of $S$ by $(u_1^a, u_2^a, \cdots, u_n^a)$ we write the well-known formula for the decomposition of the entries of a matrix

$$S_{ij} = \sum_{a=1}^{n} \lambda^a u_i^a u_j^a \quad (5)$$

so that the prescription

$$x_i^a = \sqrt{\lambda^a} u_i^a \quad (6)$$

provides the desired solution to our problem. Of course, in the case of interest $m < n$ and for a general matrix $S$, eq. (4) has no solution. A possible approach here is to look for an optimal solution in the minimum least square sense, but a far simpler and popular approach is to use the prescription (6) considering only the $m$ largest eigenvalues in the expansion (5). The quality of the approximation can then be measured by the stress function [36], [37]

$$Q = \sqrt{\frac{\Sigma_{ij}(S_{ij} - S_{ij}^*)^2}{\Sigma_{ij} S_{ij}^2}} \quad (7)$$

where $S_{ij}^* = \sum_{a=1}^{m} \lambda^a u_i^a u_j^a$ with the eigenvalues ordered such that $\lambda^1 \geq \lambda^2 \geq \cdots \geq \lambda^n$. Strictly, nowadays the designation Multidimensional Scaling Analysis is applied to the numerical minimization of $Q$ using gradient descent techniques, in which only the rank order of the entries of $S$ are used [25], [37]. Our procedure follows the original formulation of the Multidimensional Scaling proposed by [30].

The issue now is to pick a 'representative' value for the dimension $m$. A large value of $m$ yields a very low stress value (we recall that $Q = 0$ for $m = n$ by construction) but then most of the dimensions may not be relevant to describe the underlying structure of the similarity matrix as they are likely to be determined more by noise than by the essential structure of $S$. Alternatively, a too small value of $m$ may not reproduce the similarity matrix with sufficient accuracy. A popular heuristic method to determine the 'optimal' dimensionality is the so-called elbow test in which the stress $Q$ is plotted against $m$, as done in Fig. 2. Ideally, such a graph should exhibit an 'elbow' indicating that, after a certain value of $m$, the rate of decrease of the stress function becomes negligible. The results of Fig. 2 are not so discrepant from this idealistic expectation, as for $m \geq m^* \approx 30$, the dimension at which the concavity of the stress function nearly vanishes, the rate of decrease of $Q$ is of approximately 0.001, the slope of the solid straight line shown in the figure. However, there is a lot of subjectivity in the estimate of the critical dimension $m^*$ based on the elbow test, as illustrated by the two fitting straight lines in Fig. 2. In fact, the fitting of the dashed line, in which we have eliminated the first point ($m = 1$) because of its interpretation as random noise, yields a much lower estimate for this critical dimension, namely, $m^* \approx 5$.

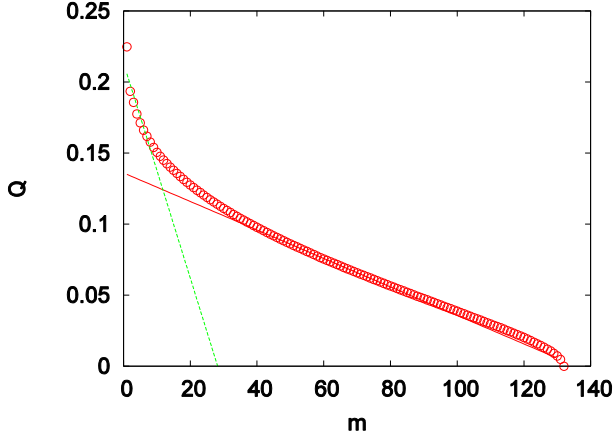

Fig. 2. Elbow test showing the stress function $Q$ as defined in Eq. (8) against the number of dimensions $m$ of the underlying word emotion space. The solid straight line has the slope $-0.001$ whereas the dashed line has slope $-0.007$.

As already pointed out, a similar mathematical analysis of the dimension of the emotion word space in which participants rated the similarity between the emotion words yielded a much smaller value for $m^*$, typically $m^* = 2$ [23] and $m^* = 3$ [29]. Of utmost interest for the Multidimensional Scaling Analysis (as well as for the Factor Analysis) is the interpretation of the eigenvectors associated to the largest eigenvalues. This type of analysis resulted in the claim that the emotion space can be described by essentially two axes (eigenvectors), namely, the degree of pleasure and the degree of arousal, and provided the main evidence in support of the Circumplex model of emotion [26]. Although this is clearly not the case here, since our critical dimension $m^*$ is definitely greater than two, it is instructive to look more closely to the first three eigenvectors of our similarity matrix $S$, shown in Fig. 3, and seek an emotional interpretation for them. We note that whereas for the main coordinate (rescaled eigenvector) $x^1$ all components are positive (upper panel of Fig. 3), all other 129 coordinates fluctuate between positive and negative values. The same behavior pattern was found in the spectral analysis of the null model similarity matrix $R$.

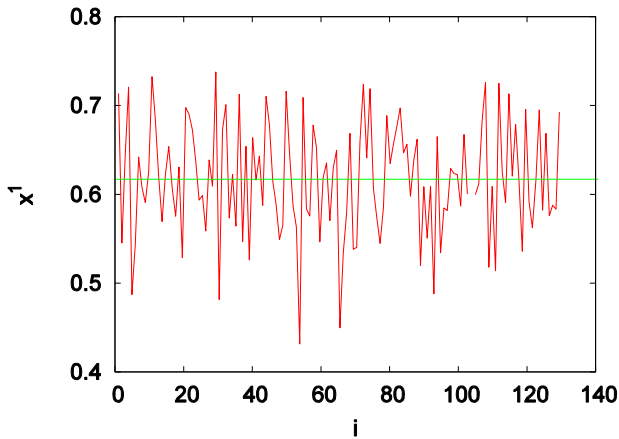

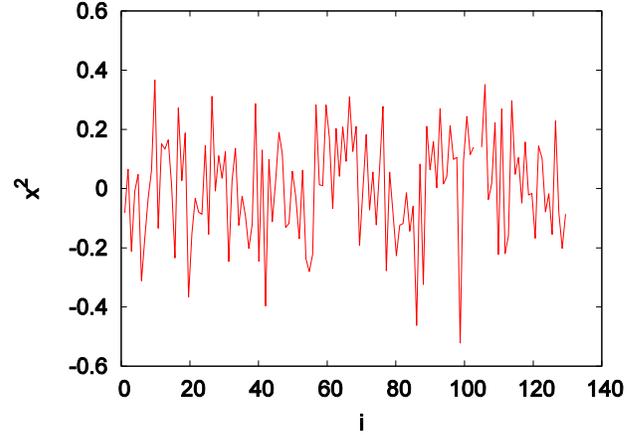

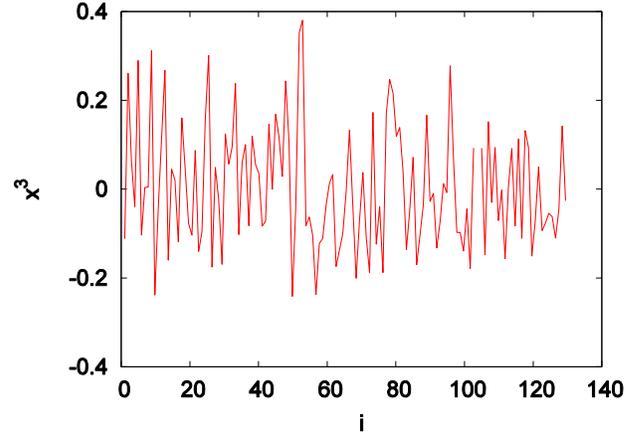

Fig. 3. The coordinate vectors $x^a$ associated to the three largest eigenvalues of the similarity matrix $S$. The labels $i = 1, \cdots, 130$ stand for the emotion words listed in Table I. The horizontal line in the upper panel indicates the mean value 0.617 of the entries of $x^1$.

To interpret the principal coordinate $x^1$, which is the first eigenvector rescaled by the square root of its corresponding eigenvalue [see eq. (6)], first we note that its eigenvalue $\lambda^1$ is about ten times greater than $\lambda^2$. We attribute such a large contribution, as well as the fact that all the entries of $x^1$ are positive, to the noisy portion of the similarity matrix $S$. In fact, the average of the entries of $x^1$, $\sum_i x_i^1/130 = 0.617$ (shown as the horizontal line in the upper panel of Fig. 3), is close to the mean of the entries of the random matrix $R$. This is expected since a great part of the similarity between any two target emotion names will be due to the coincidence between emotion unrelated words such as many classes of verbs, for instance. It is difficult to find a clear emotional interpretation of the other principal coordinates (such an interpretation is not always possible) but overall we can see that there is a psychologically meaningful contraposition between negative and positive emotion words. This is clear for the principal coordinates $x^2$ and $x^3$, which differ by the intensity of the positive and negative emotions, whose possible interpretations are, correspondingly, 'anger', and 'pleasure'.

## C. Hierarchical Clustering

The variance or spread of a set of points (i.e., the sum of the squared distances from the centre) is the key element of many clustering algorithms. In Ward's minimum variance method [31] we agglomerate two distinct clusters into a single cluster such that the within-class variance of the partition thereby obtained is minimal. Hence the method proceeds from an initial partition where all objects (130 emotion names, in our case) are isolated clusters and then begin merging the clusters so as to minimize the variance criterion.

Table II summarizes the results of the hierarchical clustering algorithm when the objects (i.e., the target emotion names) are partitioned into 25 categories, which is the highest level of hierarchy described in [29].

TABLE II

| 1 | acceptance, courage, elation, joy, sadness, sorrow, wonder |
|---|---|
| 2 | agitation |
| 3 | adoration, affection, fondness, infatuation, liking, love, lust |
| 4 | alarm, apprehension, dismay, dread, fear, fright, tenseness, uneasiness |
| 5 | aggravation, annoyance, exasperation, fury, grouchiness, grumpiness, irritation, rage, wrath |
| 6 | amusement, delight, gaiety, gladness, jolliness, joviality |
| 7 | arousal, bliss, distress, ecstasy, euphoria, hysteria, melancholy, mortification, rapture, shock, triumph |
| 8 | anger, attraction, grief, hurt, resentment, worry |
| 9 | bitterness, guilt, pity, pride, regret, remorse |
| 10 | aversion, dislike, hate |
| 11 | desire, eagerness, enthusiasm, excitement, exhilaration, jubilation, passion, zeal, zest |
| 12 | anxiety, panic |
| 13 | alienation, compassion, defeat, disappointment, frustration, insult, loneliness, pain, relief, sympathy |
| 14 | horror, terror |
| 15 | agony, anguish, gloom, glumness, misery, suffering, torment, unhappiness, woe |
| 16 | glee |
| 17 | dejection, depression, despair, hopelessness |
| 18 | anticipation, happiness, interest, optimism, satisfaction, surprise |
| 19 | embarrassment, shame |
| 20 | ferocity, outrage, spite |
| 21 | envy, insecurity, jealousy |
| 22 | contempt, disgust, loathing, revulsion, scorn, vengefulness |
| 23 | hope, humiliation |
| 24 | displeasure, hostility, neglect, pleasure |
| 25 | homesickness, isolation, rejection |

We note that although these classifications are overall reasonable there are a few pairs of antonymous words that are lumped together in the same cluster, e.g., *joy/sadness* (cluster 1) and *pleasure/displeasure* (cluster 24). At this stage, it is not clear whether this finding is the result of an imperfect context filtering scheme or whether it reflects some intrinsic property of the retrieved texts.

Some words about the clusters produced by Ward's algorithm are in order. First, as already pointed out, the initial partition contains 130 singleton clusters. The first agglomeration, which reduced the number of clusters to 129, grouped the words *aggravation* and *irritation*; the second agglomeration grouped the words *anguish* and *gloom*; the third, the words *mortification* and *shock*; the fourth, the words *eagerness* and *enthusiasm*; and the fifth, the words *euphoria* and *rapture*. Not surprisingly, these pairs of words happen to be those with the highest similarity values. At these stages of the hierarchy, the good performance of our context comparison method in clustering words of similar meanings, without employing any explicit semantic information, is truly remarkable. The consequences of this finding – a self-organized dictionary - certainly deserve further research.

## IV. CONCLUSION

In this contribution we present the first steps towards the ambitious goal of exploring the vast amount of texts readily available in the Web to obtain information about the psychology of the beings who wrote those documents. Our paper addresses the categories of English emotion names – an extensively investigated research topic in social psychology [23], [29], [32], [33], which suits very well to our research program since its staple are lexical items (i.e., a list of emotion names). Whether there is any sense in drawing inferences about emotions from people's words for the emotions is a debatable issue, which involves cross-cultural studies of emotions [38]. However, as noted by [10], the appearance of words like *angry*, *afraid*, and *happy* in all languages suggests that they represent universal experiences. In any event, words express concepts by means of which people categorize a part of their personal and social reality [38] and these categories are important constituents of people's psychology.

The findings reported in this paper were most encouraging as our proposed measure for the similarity between pairs of emotion names, which is based solely on the number of common words in the contexts of the two target emotion names, resulted in all reasonable categories that are highly correlated with the categories obtained from the subjective judgment of participants in psychology experiments [29]. Actually, our classification exhibited a few noteworthy (mis)placements of antagonistic words into the same cluster, which could easily be corrected by refinement of filtering procedure. We choose not to do so in order to exhibit this peculiar characteristic of written texts. In addition, our estimate of the dimension of the subjacent emotion space $m^*$ using the original formulation of Multidimensional Scaling [30] is not inconsistent with those obtained from people's judgment of emotion similarities. As pointed out in the discussion of Fig. 2, there is a considerable subjective component in the determination of that critical dimension.

Our approach has some advantages over the traditional method of asking participants to rate the similarity of a selection of emotion names. The most obvious one is the possibility of investigating how cultural evolution has affected our perception of emotions. In fact, studying contexts of how emotion words were used is possibly the only way to understand emotions existing centuries and millennia ago. For example, by studying usage contexts, it was suggested that even such a basic idea as 'forgiveness' in its contemporary meaning appeared only two or three centuries ago, and did not exist in antiquity, or in the Church Fathers, or in the Bible [39]. This may not be so surprising in view of the claim that Homer's characters in the Iliad and the Odyssey had no concept of 'guilt' either [40]. Another advantage of our approach is the easiness of investigating how languages and cultures differ in emotionality. Experimental studies demonstrated different emotional content in different languages [41], and it was suggested that the grammar affects the emotionality of a language [18]. The method developed here can be easily applied to different languages. Additional topics of investigation are the comparison between the categories of emotion words in prose and poetry, as well as among different writers.